\documentclass{article} 
\usepackage{iclr2023_conference,times}

\usepackage{amsmath,amsfonts,bm}

\def\eqref#1{equation~\ref{#1}}

\def\1{\bm{1}}

\DeclareMathAlphabet{\mathsfit}{\encodingdefault}{\sfdefault}{m}{sl}
\SetMathAlphabet{\mathsfit}{bold}{\encodingdefault}{\sfdefault}{bx}{n}

\usepackage{float}
\usepackage{graphicx}

\usepackage{hyperref}
\usepackage{url}
\usepackage{booktabs}

\title{Explaining Multimodal Data Fusion: \\Occlusion Analysis for Wilderness Mapping}

\author{Burak Ekim \& Michael Schmitt \ \\
Department of Aerospace Engineering\\
University of the Bundeswehr Munich\\
Neubiberg, Germany\\
\texttt{\{burak.ekim,michael.schmitt\}@unibw.de} \\
}

\iclrfinalcopy 
\begin{document}

\maketitle

\begin{abstract}
Jointly harnessing complementary features of multi-modal input data in a common latent space has been found to be beneficial long ago. However, the influence of each modality on the model's decision remains a puzzle. This study proposes a deep learning framework for the modality-level interpretation of multimodal earth observation data in an end-to-end fashion. While leveraging an explainable machine learning method, namely Occlusion Sensitivity, the proposed framework investigates the influence of modalities under an early-fusion scenario in which the modalities are fused before the learning process. We show that the task of wilderness mapping largely benefits from auxiliary data such as land cover and night time light data.

\end{abstract}

\section{Introduction}
Recent development in Earth Observation (EO) sciences enabled easier access to an abundance of EO data acquired by various sensors with different imaging techniques, opening up new venues for Deep Learning (DL) applied to EO data. Information about a phenomenon can be acquired from various types of sensors and it is challenging and almost unreal for a single modality to capture its entire rich characteristic \citet{multimodal_motivation}. Expectedly, benefiting from multimodal EO data has long been found to be more beneficial as opposed to unimodal settings in which only a single modality is used when addressing the task \citet{multimodal_review}. Enabling the modalities to interact and provide hints to each other has been found to be beneficial in many disciplines, including EO \citet{multimodal_review, fusion_review}. However, given that multimodal data do not necessarily improve performance \citet{multimoda_bad}, deciding on what modalities to use for the task at hand remains a challenge. 

Addressing this open issue, we introduce a generic end-to-end framework that enables multi-modal data investigation on the modality level. The multi-input and multi-output framework (see Figure \ref{fig:flow}) introduces the Occlusion Sensitivity Analysis \citet{zeiler_fergus} logic into the UNet \citet{unet} architecture in a modality-oriented fashion. Occlusion Sensitivity analysis is a perturbation-based method that occludes one part of the input at a time with the pixel value used for occlusion and observes the change in the output. Instead of performing standard spatial occlusion where the input image is partially occluded, we apply modality-wise occlusion with the aim of measuring the influence of a set of modalities on the models performance. The modality influence values are generated by occluding a different modality at each iteration. For each occluded modality, the model outputs an influence score that quantifies the degree to which the occluded modality contributes to the models decision. The list of modality influence values is then up-sampled and concatenated with the models output. The assumption is that the injection of modality influence values into the model uncovers hidden hints that are valuable for the supervised task of semantic segmentation. It should be noted that the proposed approach further differs from the general perspective given that it is adapted to the task of semantic segmentation. Most explainable machine learning approaches are tailored for patch-wise image classification tasks, leaving an explanation of the semantic segmentation models under-explored \citet{seggradcam}.

\section{Methodology and Experimental Setup}
\label{3}

\subsection{Methodology}
\begin{figure}[H]
\begin{center}
\includegraphics[scale=0.35]{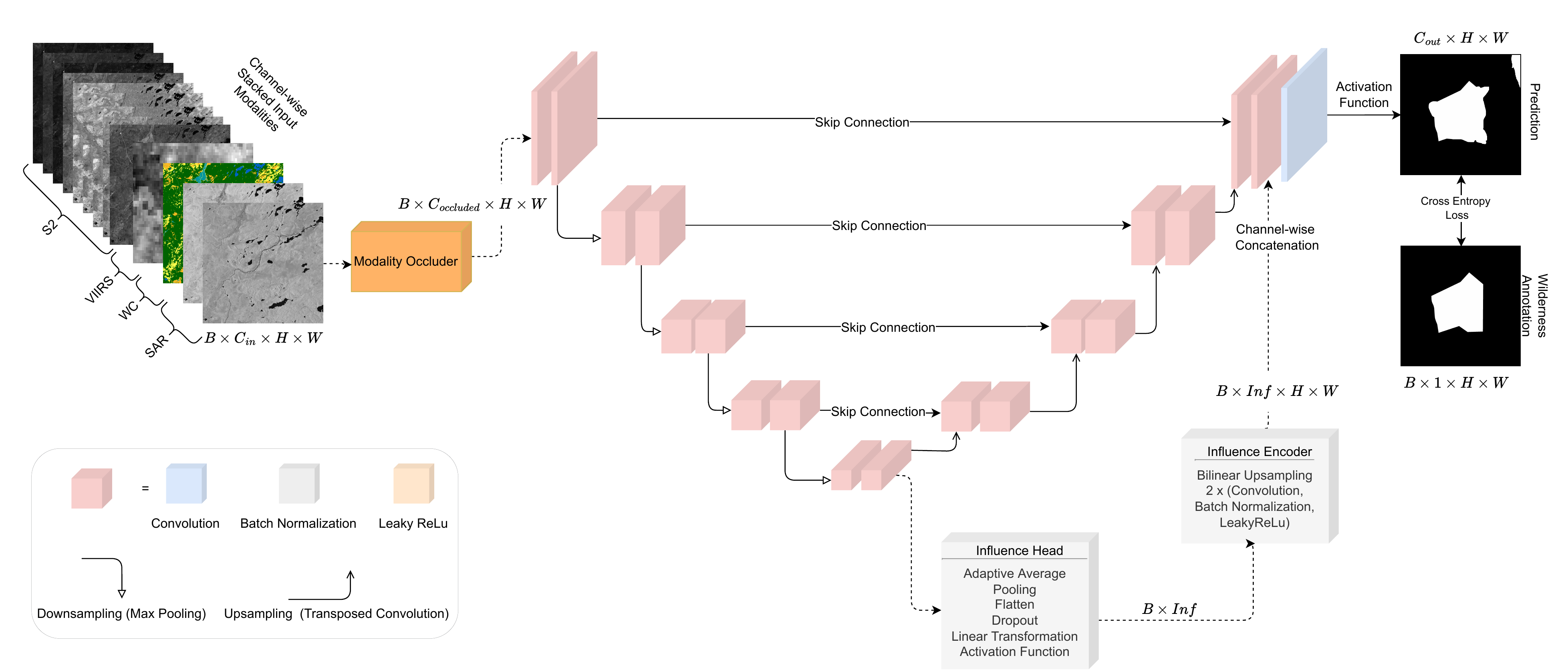}
\end{center}
\caption{The flow of the proposed framework. The dashed lines show the process of the introduced concept in which the modality influence values are quantified using the modality occluder and injected back into the model. $B$ denotes batch size, $H \times W$ denotes height and width, and $C$ denotes channels/bands (or modalities). $Inf$ is the short form for influence score. \footnotesize The illustration is created using Diagrams.net.}
\label{fig:flow}
\end{figure}

In Figure \ref{fig:flow}, the proposed logic is shown in dashed lines and solid lines indicate the flow of the plain UNet designed for the supervised semantic segmentation task. The proposed framework is built on UNet and follows a multi-modal and multi-task learning scheme. The model receives the partially occluded multi-modal image and produces two outputs (segmentation map and influence score). 

The occluded multi-modal image is created by the modality occluder (MODO) block. The MODO block, for given modality indexes, occludes the specified modalities with a specified pixel value. Alongside the output segmentation map, the model additionally outputs influence scores from the influence head that is built on top of the bottleneck layer. The influence score is an activated scalar value within the range of [0,1]. Given that the influence scores are produced for a given partially-occluded multimodal image, the score represents the contribution of the occluded modalities of the multi-modal image to the output (i.e., the higher the score, the lower the influence of the set of occluded modalities on the models decision). 

The proposed framework follows a repetitive nature. That is, for an \textit{N}-modality image where a modality-wise influence is calculated for each modality, during each pass the MODO block occludes one modality, and the influence head outputs one influence score. This process repeats itself until \textit{N} influence scores are produced, meaning each input image is passed through the model \textit{N} times. Later, the influence scores are encoded and concatenated to the last block of the model. The overall assumption is that the model could intercorporate the modality-wise influence scores in a way to value the modalities according to influence scores.  

\subsection{Dataset}
We use MapInWild \citet{miw_magazine} for the evaluation of the framework in supervised learning settings. MapInWild is a multi-modal dataset designed for the task of wilderness mapping and contains over 1000 areas that are sampled from the World Database of Protected Areas (WDPA) \citet{wdpa}, making it a suitable test-bed for the demonstration of the proposed approach. Each area in the dataset consists of a dual-pol Sentinel-1, four-season Sentinel-2, Visible Infrared Imaging Radiometer Suite (VIIRS), ESA WorldCover, and the annotations of the three WDPA classes: Strict nature reserves, wilderness areas, and national parks. 

\subsection{Experimental Setup and Results}
During the learning procedure, we use all of the modalities in MapInWild forming an image with 14 bands (10 bands Sentinel-2, 2 bands Sentinel-1, 1 band ESA WorldCover, and 1 band VIIRS). We group the bands and form the following modalities as follows: Sentinel-1: VV and VH bands; Sentinel-2(RGB): B4, B3, and B2; Sentinel-2(RGBNIR): B4, B3, B2; B8; Sentinel-2(All Bands): B2, B3, B4, B5, B6, B7, B8, B8A, B11, B12; World Cover: ESA World Cover data; Night Time Light: VIIRS Night Time Light data.

During training and validation, we apply on-the-fly cropping with a shape of $1024 \times 1024$ pixels. The batch size is set as 16 and the learning process is terminated after 6 consecutive epochs of no improvement in the observed evaluation metric. We use NVIDIA A100 80GB as a GPU. The initial learning rate is set as $10^{-3}$ and the cyclical learning rate method  that cyclically varies the learning rate between the upper limit(1) and lower limit($10^{-8}$) is used \citet{cyclic}. We use Adam with an epsilon value of $10^{-8}$ and beta values of (0.9, 0.999). The objective function measures the (Binary) Cross Entropy between the input and the target probabilities. The pixel value that is used for occluding a modality is set as 0.

The qualitative results are given in Table \ref{tab:ablation} in Intersection Over Union (IoU), accuracy, and F1 scores. 
The proposed framework introduces an occlusion-based logic to a model in an end-to-end manner and improves all of the observed evaluation metrics considerably by providing insights into the importance of input modalities. Thus, it is found that the injection of modality influence values into the model reveals hidden clues that are beneficial in the pixel-wise delineation of protected areas. 

\begin{table}[t]
\caption{Ablation Study: The plain UNet and the proposed framework on MapInWild dataset.}
\begin{center}
\begin{tabular}{l|r|r|r}
\toprule
\multicolumn{1}{c}{\bf Settings}  &\multicolumn{1}{c}{\bf IoU} &\multicolumn{1}{c}{\bf Accuracy} &\multicolumn{1}{c}{\bf F1}
 \\ 
\hline
Plain UNet         & 69.1 & 76.31 & 81.73 \\
Proposed Framework & 78.5 & 81.1 & 87.79 \\
\bottomrule
\end{tabular}
\end{center}
\label{tab:ablation}
\end{table}

\begin{table}[t]
\caption{Comparison of mean execution time in seconds: The plain UNet and the proposed framework.}
\label{wall_clock}
\begin{center}
\begin{tabular}{l|r|r|r}
\toprule
  \multicolumn{1}{c}{\bf Settings}  &\multicolumn{1}{c}{\bf Epoch} &\multicolumn{1}{c}{\bf Batch} &\multicolumn{1}{c}{\bf Optimizer Step}
\\ \hline 
Plain UNet         & 118.58 & 0.26074 & 0.25953 \\
Proposed Framework & 163.66 & 0.94645 & 0.93145 \\
\bottomrule
\end{tabular}
\end{center}    
\label{wall_clock}
\end{table}

\begin{figure}[H]
\begin{center}
\includegraphics[scale=0.20]{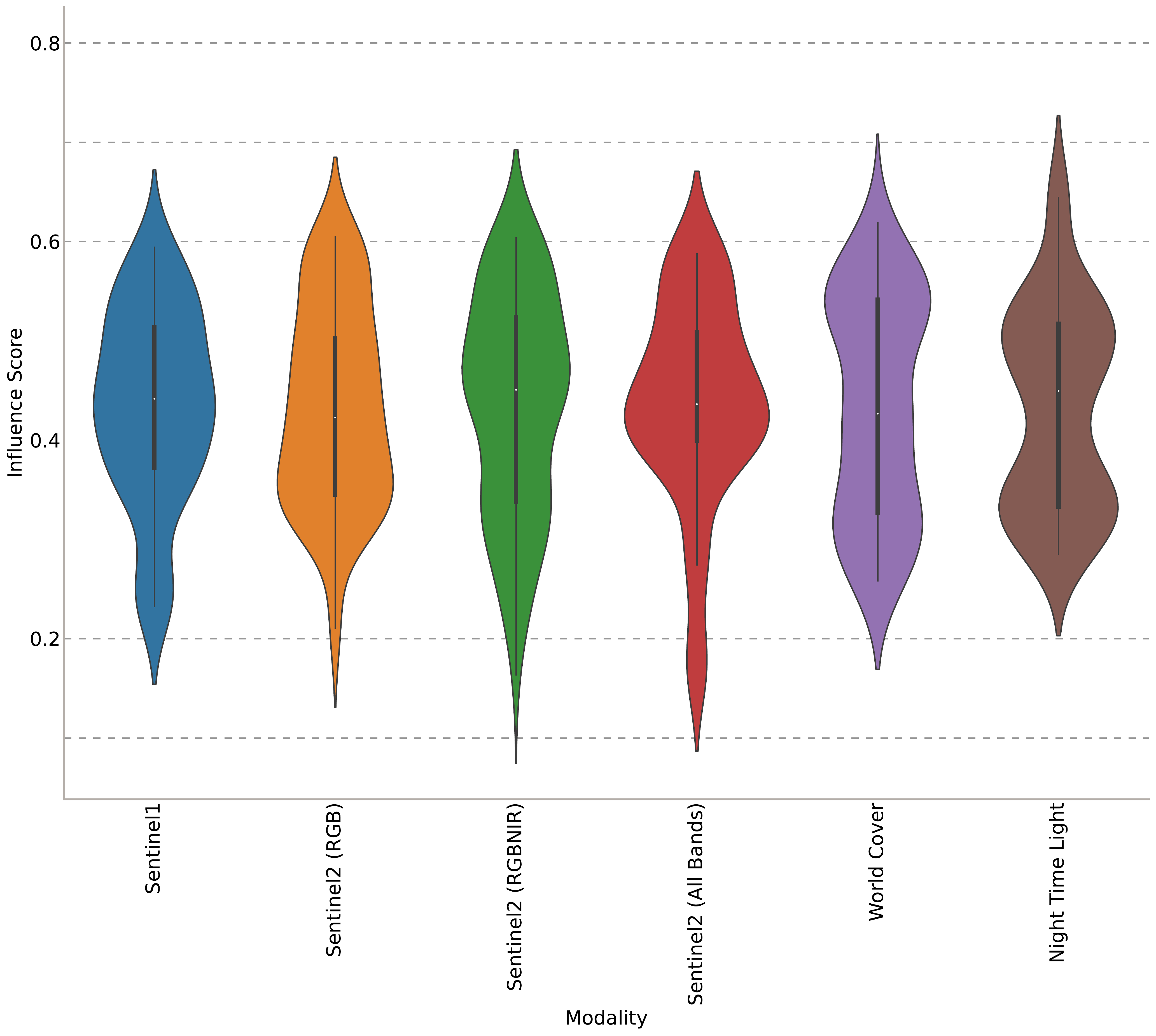}
\end{center}
\caption{The density distribution of the influence scores per modality. The distributions get wider as the number of data points lying in a specific range increases. The kernel bandwidth is computed with the \textit{silverman} method.}
\label{fig:violin}
\end{figure}

Further, in Table \ref{wall_clock} a comparison in execution time between the plain UNet and the proposed framework is reported. The proposed framework introduces an additional computational load of overall approximately 45 seconds per epoch given the observation that the model runs each batch over the model \textit{N} times, where \textit{N} is the number of modalities for which we calculate the influence scores. The violin plot is used to visualize the density distribution of the influence scores in Figure \ref{fig:violin}. The plot shows that some modalities are more influential in models decisions in classifying wilderness areas. For example, the VIIRS Night Time Light and World Cover data seem to contribute largely to the models understanding of wilderness areas. This points out that understanding wilderness areas benefits from land cover and night time illumination data. This could be explained along the lines of human activity proxy. That is, land cover and night time illumination data provide indications of surface characteristics and energy consumption. As the experiments show, these human activity indicators provide hints for better delineation of wilderness areas where human activity should be minimum. Moreover, adding more spectral information in addition to plain Sentinel-2 data was found to be beneficial (in the Figure \ref{fig:violin} notice the distribution getting more concentrated at the higher influence scores in order of $Sentinel2(RGB) \rightarrow Sentinel2(RGBNIR) \rightarrow Sentinel2(All Bands)$).

\section{Limitations and Future Work}
\label{4}
Although it seems logical in this setting for a model to understand the contribution of each modality and perform accordingly, in a future study, we intend to look into ways of enabling a higher level of interpretability concerning the models behavior in treating the modality influence values. Additionally, we plan to extend the ablation study and evaluate the proposed framework across different tasks and datasets. 

\section{Conclusion}
\label{5}
We propose a domain and multi-modal data fusion framework in which an explainable machine learning technique is embedded in the model in an end-to-end fashion.  
Shortly, we reformulate the occlusion logic and perform channel-level sensitivity analysis for the specific task of wilderness mapping from EO data. Given the observation that the modality influence values provide valuable information for the model to better perform, the proposed logic could be seen as a realization of a model that guides itself in learning the influence of each modality for the task at hand. We believe this approach could be of great use for the EO research where access to multimodal data is easy.
\bibliography{iclr2023_conference}
\bibliographystyle{iclr2023_conference}

\end{document}